\documentclass[letterpaper, 10 pt, conference]{styles/ieeeconf}  

\IEEEoverridecommandlockouts                              

\usepackage{tikz}
\usepackage{booktabs, multirow}
\usetikzlibrary{arrows.meta, positioning, calc}

\usepackage{amsmath,amsfonts,bm,amssymb}
\usepackage{mathtools}

\def\eqref#1{equation~\ref{#1}}

\def\1{\bm{1}}

\DeclareMathAlphabet{\mathsfit}{\encodingdefault}{\sfdefault}{m}{sl}
\SetMathAlphabet{\mathsfit}{bold}{\encodingdefault}{\sfdefault}{bx}{n}

\DeclareMathOperator*{\argmin}{arg\,min}

\usepackage[acronym]{glossaries}
\newacronym{dof}{DOF}{degrees of freedom}
\newacronym{apf}{APF}{artificial potential fields}
\newacronym{osc}{OSC}{operational space control}
\newacronym{cbf}{CBF}{control barrier functions}

\usepackage{tikz}
\usepackage{booktabs, multirow}
\usetikzlibrary{arrows.meta, positioning, calc}
\usepackage{subcaption}
\usepackage{hyperref}

\let\ACMmaketitle=\maketitle
\renewcommand{\maketitle}{\begingroup\let\footnote=\thanks \ACMmaketitle\endgroup}

\makeatletter
\newcommand*\titleheader[1]{\begingroup\gdef\@titleheader{#1}\let\footnote=\thanks\endgroup}
\AtBeginDocument{%
  \let\st@red@title\@title
  \def\@title{%
  \begin{flushleft}
    \vspace{-2.0em}
    \bgroup\normalfont\small\@titleheader\par\egroup
    \vspace{-18pt}\par\noindent\rule{\textwidth}{0.1pt}
    \end{flushleft}
    \vskip0.5em\st@red@title
        }

}

\title{\LARGE \bf
A Safety-Aware Shared Autonomy Framework with BarrierIK Using Control Barrier Functions
}

 \author{
  Berk Guler$^{\star1,6}$, Kay Pompetzki$^{\star1}$, Yuanzheng Sun$^{1}$, Simon Manschitz$^{6}$, Jan Peters$^{1-5}$
\thanks{$^\star$Equal contribution; $^1$Technical University of Darmstadt $^2$German Research Center for Artificial Intelligence (DFKI) $^3$hessian.AI $^4$Robotics Institute Germany (RIG) $^5$Centre for Cognitive Science $^6$Honda Research Institute Europe GmbH; Corresponding author: \texttt{berk.gueler@tu-darmstadt.de}}
\thanks{This work was supported by the German Federal Ministry of Research, Technology and Space (BMFTR) under the Robotics Institute Germany (RIG) and the Honda Research Institute EU, Germany, and the EU’s Horizon Europe project ARISE (Grant no.: 101135959).}
\thanks{This work has been accepted for publication in the Proceedings of the 2026 IEEE International Conference on Robotics and Automation (ICRA 2026). The final published version will be available via IEEE Xplore.}%
}
\titleheader{\textit{Preprint}}
\begin{document}

\maketitle
\thispagestyle{empty}
\pagestyle{empty}

\begin{abstract}
Shared autonomy blends operator intent with autonomous assistance. In cluttered environments, linear blending can produce unsafe commands even when each source is individually collision-free. Many existing approaches model obstacle avoidance through potentials or cost terms, which only enforce safety as a soft constraint. In contrast, safety-critical control requires hard guarantees. We investigate the use of control barrier functions (CBFs) at the inverse kinematics (IK) layer of shared autonomy, targeting post-blend safety while preserving task performance.  Our approach is evaluated in simulation on representative cluttered environments and in a VR teleoperation study comparing pure teleoperation with shared autonomy. Across conditions, employing CBFs at the IK layer reduces violation time and increases minimum clearance while maintaining task performance. In the user study, participants reported higher perceived safety and trust, lower interference, and an overall preference for shared autonomy with our safety filter. Additional materials available at \href{https://berkguler.github.io/barrierik/}{BarrierIK}.
\end{abstract}

\section{Introduction}
\label{sec:introduction}
In teleoperation, shared autonomy combines operator input with autonomous assistance to improve task performance~\cite{selvaggio2021autonomy}. Two common approaches are (i) blending methods, where operator and autonomy commands are linearly blended against each other~\cite{dragan2013policy,muelling2017autonomy,jeon2020shared,gopinath2016human,song_2024_robot,atan_2025_sa}, and (ii) policy methods, where operator input is treated as an observation to weight among predefined or learned assistance policies~\cite{javdani2015shared,javdani2018shared,nikolaidis2017human,khoramshahi2019dynamical,allenspach_2024_task}. These paradigms improve efficiency and reduce workload, but they typically arbitrate in task space without enforcing constraints. As a result, even if each command is individually safe, the blended action may still violate safety requirements such as collision avoidance~\cite{dragan2013policy,gopinath2016human,jeon2020shared, gottardi_2022_sc_potential_fields,song_2024_robot}.

Several approaches have been proposed to address constraint satisfaction in teleoperation. Many integrate collision avoidance into the objective function, for example, through artificial potential fields~\cite{khatib1986real,khatib2003osc,nakanishi2008operational,ratliff2018riemannian,cheng2020rmp} or cost terms. Others enforce feasibility by filtering operator commands through collision-aware inverse kinematics (IK)~\cite{rakita_2021_collisionik,wang2023rangedik}. While successful, these approaches treat safety as a soft constraint, which allows violations when task objectives dominate. In addition, modeling safety as an extra cost term can lead to conflicting behaviors and carries the risk of lowering user satisfaction.

\begin{figure}[t!]
    \centering
    \includegraphics[width=\linewidth]{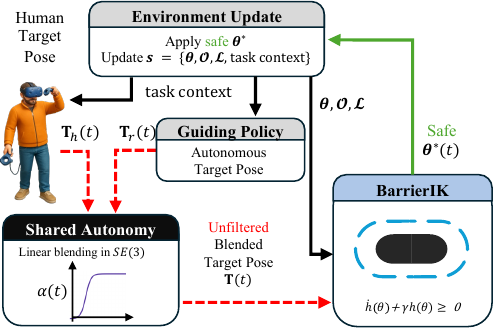}
\vspace{-1.5em}
    \caption{
Pipeline with CBF-based safety filtering \emph{after} arbitration. The operator specifies a target pose~$\mathbf{T}_{\mathrm{h}}(t)$; the guiding policy outputs~$\mathbf{T}_{\mathrm{r}}(t)$ from the current state $\boldsymbol{s}(t)~=~\{\boldsymbol{\theta}(t), \boldsymbol{\mathcal{O}}(t), \boldsymbol{\mathcal{L}}(t), \text{task context}\}$, where $\boldsymbol{\theta}$ is the joint configuration, $\boldsymbol{\mathcal{O}}$ and $\boldsymbol{\mathcal{L}}$ are obstacle and robot link capsules, and ``task context'' includes task-specific info (e.g., rendered views). These unfiltered poses are blended in $SE(3)$ via arbitration weight~$\alpha(t)$ to form~$\mathbf{T}(t)$ (red dashed), which may violate safety margins. \textit{BarrierIK} receives $\mathbf{T}(t)$ and enforces CBF conditions to compute a safe joint command~$\boldsymbol{\theta}^{*}(t)$ (green), which is applied to the robot. The updated state~$\boldsymbol{s}(t)$ is fed back to the policy and user.
}


    \label{fig:experiment:hierarchy}
    \vspace{-2em}
\end{figure}
A promising alternative is the use of control barrier functions (CBFs)~\cite{ames_2014_control,ames_2016_control,ames_2019_control}. CBFs are designed to ensure forward invariance of safety sets, offering a framework for formal constraints such as collision avoidance. Unlike cost-based methods, they do not compromise between safety and task performance. Instead, they morph potentially unsafe commands onto the safe set via constrained quadratic programming. CBFs have been successfully applied and teleoperation~\cite{xu_2018_cbf,zhang2021cbfteleop,zhang2020,qin2025hapticcbf}, but their application in shared control and shared autonomy remains limited~\cite{he2021barrier}. Advances in operational-space formulations now provide efficient scalability~\cite{oscbf}, motivating our approach to strictly enforce feasibility and aim for collision-free motion in high-DOF manipulation tasks.

We therefore propose BarrierIK, a CBF-constrained inverse-kinematics solver placed after policy blending. We hypothesize that projecting blended commands into the safety set, rather than treating safety as a soft objective, preserves task performance while increasing constraint satisfaction and maintaining user satisfaction. The pipeline follows the shared-autonomy order: (i) blend the task-space command, then (ii) target safety at the IK layer. We evaluate BarrierIK in simulation
on cluttered manipulation scenarios with dynamic and static obstacles, as well as in a VR teleoperation study with ten participants. 
In all tested scenarios, BarrierIK reduces violation time and increases clearance while maintaining task performance. The user study further indicates a consistent trend supporting our hypothesis: BarrierIK as a safety filter improves the safety and keeps performance in shared autonomy while preserving user satisfaction.

This paper makes three contributions. First, we present a shared-autonomy architecture that blends task-space commands and then addresses post-blend safety at the IK layer, handling failure modes of command-space blending in cluttered, nonconvex scenes. Second, we introduce BarrierIK, a CBF-constrained IK formulation that treats safety as a hard inequality constraint while remaining compatible with standard IK objectives, including pose tracking, joint limits, and smoothness. Third, we provide a comparative evaluation in simulation (dynamic obstacles and frame/shelf) and in a VR teleoperation study with ten participants, contrasting the proposed approach with IK baselines and quantifying benefits for both safety and task performance.
	

\section{Related Work}
\label{sec:related_work}
In this section, we review three areas relevant to our scope. First, we discuss shared autonomy methods that incorporate autonomous assistance in teleoperation. Second, we address safety with a focus on collision avoidance. Finally, we review control barrier functions (CBFs), which provide formal safety guarantees as well as a principled way to incorporate safety.
\paragraph*{\textbf{Shared Autonomy (SA)}} aims to enhance operator performance and reduce workload by incorporating autonomous guidance~\cite{selvaggio2021autonomy}. A common paradigm is linear policy blending, where the final action is a linear combination of the operator's and autonomy's commands, based on inferred user intent~\cite{dragan2013policy, gopinath2016human, muelling2017autonomy, jeon2020shared, gottardi_2022_sc_potential_fields, atan_2025_sa}. The autonomy policy typically take the form of optimization problems~\cite{dragan2013policy, muelling2017autonomy, gottardi_2022_sc_potential_fields}, movement primitives~\cite{khoramshahi2019dynamical, maeda2022blending}, or encode the behavior from data~\cite{jeon2020shared, song_2024_robot, atan_2025_sa}. The arbitration value controlling the trade-off between operator and autonomy commands depends on the confidence of the goal probability estimate, which is usually derived from the maximum entropy principle~\cite{dragan2013policy, muelling2017autonomy, gottardi_2022_sc_potential_fields} or recursive Bayesian inference~\cite{jain_2019_probabilistic, atan_2025_sa}.

Other works focus on policy-based methods, where the operator's command is treated as an observation that influences predefined or learned assistance policies~\cite{javdani2015shared, javdani2018shared, nikolaidis2017human, allenspach_2024_task}. These approaches formalize shared autonomy as partially observable Markov decision processes (POMDPs) for goal-aware action selection~\cite{javdani2015shared, javdani2018shared, nikolaidis2017human}, or arbitrate over movement primitives based on operator commands~\cite{ewerton2020assisted, allenspach_2024_task}. Cost functions for these POMDP formulations and movement primitives are typically predefined or learned from data.

Although many arbitration mechanisms exist, linear policy blending remains one of the most widely used methods. However, it is prone to failure in cluttered or occluded scenes: even when human and autonomy commands are individually safe, their combination may fall outside a nonconvex safe set, leading to unsafe or misaligned actions~\cite{trautmancybernetics}. These conflicts are exacerbated when multimodality in the joint human–robot distribution is ignored. A natural extension of linear blending to incorporate safety is the use of attracting and repelling fields~\cite{khatib1986real}, which add safety considerations but still treat safety as a soft constraint~\cite{jeon2020shared, gottardi_2022_sc_potential_fields, song_2024_robot}. In navigation-style SA, goal-level blending arbitrates at the goal/distribution level instead of directly in command space, improving responsiveness in clutter~\cite{Jiang_Warnell_Stone_2021}. Human factors results also show that making assistance legible (e.g., via haptic or transparent SA) improves agreement and satisfaction compared to opaque blends~\cite{DBLP:journals/corr/abs-2012-04283}.

In this work, we retain the standard SA pipeline (blending first) but place safety enforcement after the blend at the IK layer. This ensures that post-blend actions prioritize safety and collision-avoidance while remaining consistent with the operator's intent.

\paragraph*{\textbf{Safety in teleoperation}} spans multiple notions beyond geometry, including limiting interaction forces and ensuring closed-loop stability~(e.g., time-domain passivity), and active constraints/virtual fixtures that restrict motion within task-space envelopes~\cite{bowyer2014activeconstraints,ryu2004tdpa}. Here we focus on \emph{environment collision safety} for the manipulator during assisted operation.

\paragraph*{Collision avoidance in teleoperation}
For teleoperated manipulation in \emph{cluttered} environments, collision avoidance is commonly handled at the IK layer. Beyond IK, sampling-based teleoperation optimizers generate collision-aware updates by selecting feasible joint states near the current pose in real time~\cite{manschitz2025sampling}. In this section, we group IK layer methods into (i) optimization-based solvers and (ii) differential/null-space approaches. A canonical null-space strategy leverages operational space control~\cite{khatib2003osc,nakanishi2008operational} and projects an obstacle-avoidance gradient into the Jacobian's null space so the end-effector task is preserved while shaping self-motion~\cite{nullspace}. However, such methods require robot redundancy and are limited by their hierarchical nature, where one task (e.g., end-effector motion) must be prioritized over others (e.g., collision avoidance).

Among optimization-based IK methods, CollisionIK augments a multi-objective IK with environment-distance penalties and evaluates a pruned active set of nearest link–obstacle pairs each step for tractability~\cite{rakita_2021_collisionik}. This yields smooth objectives and real-time rates, but safety remains \emph{tradeable}: in narrow passages, the solver can accept transient violations to reduce tracking error, and behavior depends on active-pair selection (and weight tuning). Hard-constraint variants enforce minimum-distance inequalities directly in QP-based IK solver and report real-time collision-free updates when constraints are feasible~\cite{ashkanazy2023ikinqp}. These turn safety into a constraint (not a cost) but require reliable signed distances and careful aggregation across many contacts. When the task is infeasible at the chosen margin, they must either relax constraints or deviate substantially from the user's command. Differential IK methods that do not rely on null space have also been proposed for tight, cluttered workspaces (e.g., DawnIK~\cite{marangoz2023dawnik}); their myopic linearization supports high-rate control but offers no forward-invariance guarantee and can be sensitive to step-size/regularization choices. We therefore adopt an optimization-based IK formulation to support both redundant and non-redundant manipulators.

\paragraph*{\textbf{Control barrier functions (CBFs)}}
provide sufficient conditions for forward invariance of a safe set and are typically enforced via minimal-deviation quadratic programs~\cite{ames_2014_control,ames_2016_control,ames2019cbf}. In teleoperation, CBF filters have been used to guarantee geometric and interaction constraints while staying close to user input~\cite{xu_2018_cbf,zhang2021cbfteleop,zhang2020,qin2025hapticcbf}. 
For example, QP filters are employed to constrain interaction forces in robotic surgery ~\cite{qin2025hapticcbf}. 
In shared autonomy, works such as~\cite{he2021barrier} address safety using barrier pairs, but are primarily focused on low-dimensional robotic navigation tasks. 
Recent developments extend CBFs to task-consistent operational-space control, providing kinematic and dynamic formulations that scale to hundreds or even thousands of simultaneous safety constraints~\cite{oscbf}. These advancements motivate our work, where we introduce a post-blend safety filter that intercepts the blended user-autonomy command and resolves it at the IK layer. In this approach, we retain an optimization-based IK formulation while enforcing safety through CBF-style inequalities, treating safety as a hard inequality constraint after blending.


\section{A Safety Aware Shared Autonomy Framework}
\label{sec:methods}
In this section, we detail our real-time teleoperation framework. 
The system operates in two stages, as depicted in \hyperref[fig:experiment:hierarchy]{Figure~\ref*{fig:experiment:hierarchy}}: a high-level policy blending module combines user and autonomous motion commands via linear interpolation to generate a nominal Cartesian target. 
This command is then passed to an IK solver that ensures feasibility and safety with respect to kinematic and environmental constraints. 
We describe the policy blending strategy in \hyperref[subsec:blending]{Section~\ref*{subsec:blending}}, followed by details on three IK solvers—Baseline N~(N), Baseline P~(P), and our novel Control Barrier Function-based BarrierIK (B).

\paragraph*{\textbf{Shared Autonomy}}
\label{subsec:blending}
We assume the human operator provides Cartesian pose commands at each time step~$t$, represented as a transformation~$\mathbf{T}_\text{h}(t) \in SE(3)$, consisting of a position~$\mathbf{x}_\text{h}(t)~\in~\mathbb{R}^3$ and an orientation~$\mathbf{q}_\text{h}(t) \in \mathbb{H}$ (a unit quaternion). 
Simultaneously, the autonomous system proposes a reference pose~$\mathbf{T}_\text{r}(t) = (\mathbf{x}_\text{r}(t), \mathbf{q}_\text{r}(t))$, derived from a grasp planner. 
In our experiments, these poses are heuristically defined based on geometric primitives associated with the target objects; however, the framework is agnostic to the grasp planner and can accommodate more sophisticated methods.

The final blended pose~$\mathbf{T}(t) = (\mathbf{x}(t), \mathbf{q}(t))$ is computed by interpolating both position and orientation. 
The position is blended linearly
$
{\mathbf{x}(t) = (1 - \alpha_t)\, \mathbf{x}_\text{h}(t) + \alpha_t\, \mathbf{x}_\text{r}(t)}
$,
where~$\alpha_t \in [0, 1]$ is the arbitration coefficient that modulates the influence of the autonomous system. 
The orientation is blended using spherical linear interpolation~(SLERP)
$
{\mathbf{q}(t) = \text{slerp}\left(\mathbf{q}_\text{h}(t), \mathbf{q}_\text{r}(t), \alpha_t\right).}
$,
To avoid discontinuities caused by antipodal quaternions, we enforce a positive dot product between~$\mathbf{q}_\text{h}(t)$ and~$\mathbf{q}_\text{r}(t)$ prior to interpolation. 
The resulting blended pose~$\mathbf{T}(t)$ is passed to the downstream IK solvers to generate a safe and executable robot configuration.

The arbitration coefficient~$\alpha_t$ can be fixed or adapted dynamically based on the disagreement between the user and robot commands. 
Following works like \cite{muelling2017autonomy}, we employ a sigmoid function for ~$\alpha_t$ over the position-space disagreement
\begin{align*}
\textstyle \alpha_t = \textstyle\sigma\left(p \left( \frac{\|\mathbf{x}_\text{h}(t) - \mathbf{x}_\text{r}(t)\|_2}{s} + b \right)\right), 
\end{align*}
where~$\sigma(\cdot)$ denotes the logistic function,~$p$ is a slope parameter,~$s$ is a scaling factor, and~$b$ is a bias term. 
This arbitration strategy reduces autonomy when human and robot intent diverge, and increases it when they align—encouraging fluid collaboration. 
The arbitration function follows the approach introduced by Muelling et al.~\cite{muelling2017autonomy}.

\paragraph*{\textbf{Inverse Kinematics}}
\label{subsec:ik}
Given a desired end–effector pose $\mathbf{T}(t)\!\in\!SE(3)$, we implemented a non-linear,optimization-based inverse kinematics (IK) solver to compute joint configurations~$\bm {\theta} \in \mathbb{R}^n$, where~$n$ is the number of degrees of freedom~(DoF).
Our implementation builds on prior work~\cite{rakita_2018_relaxedik}, which enables smooth, feasible motions while accounting for pose tracking and self-collision avoidance by ensuring smoothness of motion.
We re-implemented this framework in JAX to enable high-performance automatic differentiation for forward kinematics, self-collision checking, and differentiable collision detection algorithms for the IK solvers~\cite{jax2018github, ferretti_accelerated_optimization_2025, tracy2022diffpillsdifferentiablecollisiondetection}.

The IK problem is posed as a constrained optimization
\begin{equation}
\begin{aligned}
\textstyle
\bm{\theta}^* &= \textstyle\argmin_{\bm{\theta}} \; \mathcal{J}(\bm{\theta}) \\
\text{s.t.} ~~ & 
\textstyle c_k(\bm{\theta}) \leq 0,~k = 1, \ldots, K,~~ l_i \leq \theta_i \leq u_i ~\forall i.
\end{aligned}
\label{eq:generic_opt_formula}
\end{equation}

Here, $c_k(\bm{\theta})$ are inequality constraints~(e.g., for manipulability or CBF constraint for BarrierIK), and $[l_i, u_i]$ denote the joint limits. 
The objective function~$\mathcal{J}(\bm{\theta})$ is defined as a weighted sum of differentiable task-specific terms
$
{\mathcal{J}(\bm{\theta}) = \sum_{m=1}^{M} w_m \, g_m(\mathbb{z}_m(\bm{\theta}))}$,
where $w_m \in \mathbb{R}^{+}$ is the weight, $\mathbb{z}_m(\bm{\theta})$ is a differentiable task feature, and $g_m(\cdot)$ is a scalar penalty function of the $m$-th objective. We solve~\eqref{eq:generic_opt_formula} using Sequential Least Squares Programming (SLSQP), an SQP method that, at each iteration, linearizes the constraints and minimizes a quadratic model of the Lagrangian, yielding a sequence of QP subproblems~\cite{SLSQP}.
Because SLSQP relies on local linearizations, it enforces constraint feasibility only up to a numerical tolerance, providing approximate rather than continuous-time safety guarantees.

Following \cite{rakita_2018_relaxedik}, the task features $\mathbb{z}(\bm{\theta})$ includes (i) end-effector tracking that penalizes position and orientation errors relative to $\mathbf{T}(t)$, (ii) motion-smoothness regularizers on joint velocities/accelerations/jerks and on Cartesian velocity, and (iii) a self-collision avoidance term that penalizes proximity between robot links using an ANN predictor of link-pair distances conditioned on $\bm{\theta}$. In our JAX implementation, this network remains fully differentiable through the optimizer steps. 
In addition, we impose a manipulability constraint $c_m(\bm{\theta})$ that keeps the Jacobian well-conditioned by preventing the smallest singular value from falling below a small threshold and capping the condition number at each step.

We use this IK solver in the following sections as a baseline, referred to as \textbf{Baseline~N}. It includes the above-mentioned objectives and constraints to comprise the tracking, smoothness, and self-collision avoidance together with the manipulability constraint, but does not model obstacles.

\paragraph*{\textbf{BarrierIK: Control Barrier Function-Based Environment-Aware Inverse Kinematics}}
\label{subsubsec:bik}

We introduce \textit{BarrierIK}, an IK formulation based on Control Barrier Functions (CBFs), to target safety during teleoperation, by preserving the objectives and manipulability constraints as defined in Baseline N.

For each obstacle $o\!\in\!\mathcal{O}$ we define the control barrier function ${h_o(\bm{\theta})=\phi_o(\bm{\theta})-\epsilon}$ where $\phi_o$ is the minimum signed robot–obstacle distance and $\epsilon$ is the safety margin. 
In practice, we identify the active link–obstacle pair that realizes $\phi_o$ (i.e., the closest pair) and compute its gradient via automatic differentiation \cite{tracy2022diffpillsdifferentiablecollisiondetection}. 

We enforce a discrete-time CBF condition of the form
$
{\nabla h_o(\bm{\theta})^\top \Delta\bm{\theta} + \mathcal{K}\!\big(h_o(\bm{\theta})\big) \;\ge\; 0},
$
with extended class-$\mathcal{K}$ function $\mathcal{K}(h)=\gamma h+\beta h^3$ where  $\gamma, \beta > 0$ are class-$\mathcal{K}$ coefficients controlling convergence to the safe set boundary. This formulation adopts a position-based update rule~($\Delta\bm{\theta}$) rather than a velocity-controlled formulation; while this deviates from classical continuous-time CBFs, such a discretized version suits our position-based control loop as a design choice.

To avoid discontinuities in constraint switching and to mitigate underflow or overflow issues, we aggregate all CBF constraints using a temperature-scaled soft-max (log-sum-exp) approximation
\begin{align*}
 \textstyle
c_{\mathrm{CBF}}(\bm{\theta}) = \tfrac{1}{T}\,\log\!\sum_{o\in\mathcal{O}}\exp\!\Big(T\big[-\,\dot{h}_o(\bm{\theta})-\mathcal{K}(h_o(\bm{\theta}))\big]\Big),   
\end{align*}
where $\dot{h}_o(\bm{\theta}) = \nabla h_o(\bm{\theta})^\top \Delta \bm{\theta}$. The temperature parameter $T$ controls the max-approximation (higher $T\!\Rightarrow\!$ closer to $\max$).

The optimization problem of the BarrierIK

\begin{equation}
\begin{aligned}
\textstyle
\bm{\theta}^* &= \textstyle\argmin_{\bm{\theta}} \; \mathcal{J}(\bm{\theta}) \\
\text{s.t.} ~~ & 
\textstyle c_m(\bm{\theta}) \leq 0,~~
c_{\mathrm{CBF}}(\bm{\theta})\le 0,~~
l_i \leq \theta_i \leq u_i~\forall i,
\end{aligned}
\label{eq:generic_opt_formula}
\end{equation}

minimizes the same objective and retains the manipulability constraints as defined in Baseline~N, along with the additional control–barrier inequality $c_{\mathrm{CBF}}(\bm{\theta})\le0$. The CBF constraint enforces a stepwise safety margin while tracking and promotes forward invariance of the safe set under sufficiently small updates.


Instead of imposing obstacle avoidance as a hard constraint in BarrierIK, we employ it as a soft constraint inside the objective referred to as \textbf{Baseline P}. Following the proximity-penalty formulation in~\cite{rakita_2021_collisionik}, it incorporates environment awareness by penalizing proximity between the robot and surrounding obstacles.
For each obstacle~$o \in \mathcal{O}$ and each robot link~$\ell \in \mathcal{L}$, we compute the signed distance~$\phi_{\ell o}(\bm{\theta})$ between the link and the obstacle, and define the corresponding collision objective as
$\chi_{\ell o}(\bm{\theta}) = w_{\textrm{safe}}(\phi_{\ell o}^2(\bm{\theta}) + \delta)^{-1}$,
where $w_{\text{safe}} = (5\epsilon)^2$ is a scaling factor based on the scalar safety margin~$\epsilon$, and $\delta$ is a small regularization constant to avoid division by zero.
The final objective of Baseline P
\begin{equation*}
\textstyle
\mathcal{J}_{\text{P}}(\bm{\theta}) = \mathcal{J}_{\text{N}}(\bm{\theta}) + w_{\text{col}} \cdot g_{\text{col}}(\mathbb{z}_{\text{col}}(\bm{\theta})),
\end{equation*}
augments the Baseline N objective~$\mathcal{J}_{\text{N}}(\bm{\theta})$, which combines end-effector tracking, motion smoothness, and self-collision avoidance by adding a collision penalty over all link–obstacle pairs $
{\mathbb{z}_{\text{col}}(\bm{\theta}) = \sum_{\ell \in \mathcal{L}} \sum_{o \in \mathcal{O}} \chi_{\ell o}(\bm{\theta})}$,
where each $\chi_{\ell o}(\bm{\theta})$ penalizes proximity between link $\ell$ and obstacle $o$. Also, manipulability constrained preserved $c_{\mathrm{m}}$ as defined in \hyperref[subsec:ik]{Section~\ref*{subsec:ik}}.

The scalar weight~$w_{\text{col}}$ controls the relative importance of obstacle avoidance in the overall cost. This soft-penalty formulation allows the robot to trade off safety against task objectives in a smooth and differentiable manner, but it does not guarantee forward invariance or strict constraint satisfaction.


\section{Experiments}
\label{sec:experimental-setup}
\begin{figure}[t!]
    \centering
    \includegraphics[width=\linewidth]{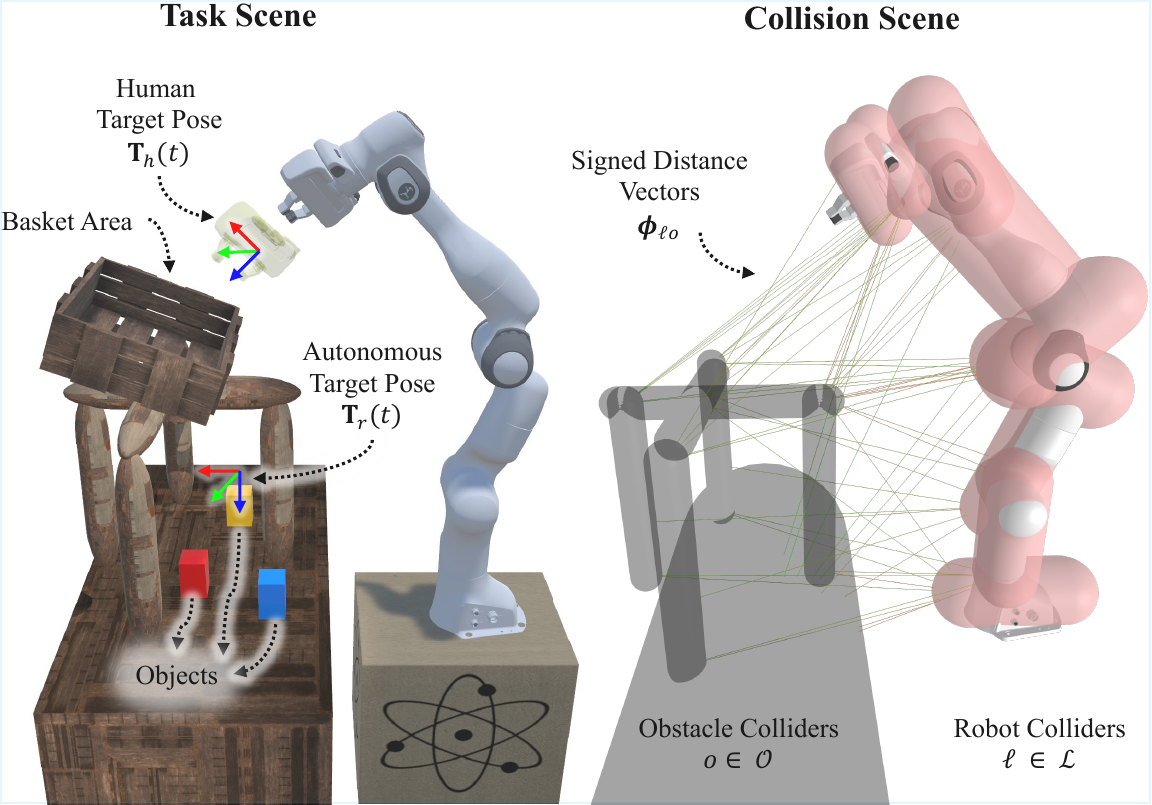}
    \vspace{-1.5em}
    \caption{
    Simulated task and collision-evaluation setup.
    \textbf{Left:} scene with human target $\mathbf{T}_\text{h}(t)$ and autonomous reference $\mathbf{T}_\text{r}(t)$.
    \textbf{Right:} robot link (red) and obstacle (black) convex-capsule colliders; for each link $\ell\in\mathcal{L}$ and obstacle $o\in\mathcal{O}$ we compute signed distances $\phi_{\ell o}(\bm{\theta})$ and signed distance vectors used by Baseline~P and BarrierIK (colors encode magnitude).
    }
    
    \label{fig:introduction}
    \vspace{-2em}
\end{figure}

We evaluate the CBF-based safety layer in autonomous mode and teleoperation to test whether \textbf{BarrierIK} improves safety, performance, and user experience over two baselines: \textbf{N}~(no environment collision handling) and \textbf{P}~(soft penalty-based collision avoidance). We report collisions, clearance, tracking, and smoothness, marking significant B vs.\ P differences. To model the environment and robot links, we adopt a differentiable collision checking algorithm for capsule primitives~\cite{tracy2022diffpillsdifferentiablecollisiondetection}, as illustrated in Figure~\ref{fig:introduction}.

\begin{table*}[t]
\centering
\footnotesize
\setlength{\tabcolsep}{2pt}
\caption{Autonomous trajectory evaluation (mean $\,\pm\,$ s.d.). For \emph{Min. clearance}, higher is better (negative $\Rightarrow$ penetration). For all other metrics, lower is better. \textbf{Bold} indicates a statistically significant B vs.\ P difference (N serves only as a reference; excluded from the significance analysis test).}
\label{tab:auto-traj}
\begin{tabular}{llrrrrrrr}
\toprule
\textbf{Task} & \textbf{Solver} & \textbf{\#Collisions} & \textbf{Min. clearance [m]} & \textbf{Violation time [\%]} & \textbf{Pos. err. [m]} & \textbf{Ori. err. [$^\circ$]} & \textbf{Task jerk [m/s$^3$]} & \textbf{Joint jerk [rad/s$^3$]} \\
\midrule
\multirow{3}{*}{Shelf}
 & N & 44.13 $\,\pm\,$ 9.74 & -0.162 $\,\pm\,$ 0.016 & 24.52 $\,\pm\,$ 1.19 & 0.020 $\,\pm\,$ 0.001 & 4.15 $\,\pm\,$ 0.07 & 4.197 $\,\pm\,$ 0.354 & 48.15 $\,\pm\,$ 2.89 \\

 & P & 19.30 $\,\pm\,$ 5.85 & -0.171 $\,\pm\,$ 0.017 & 4.26 $\,\pm\,$ 5.22 & 0.049 $\,\pm\,$ 0.018 & \textbf{4.50 $\,\pm\,$ 0.12 }& 6.518 $\,\pm\,$ 0.738 & \textbf{57.22 $\,\pm\,$ 3.35} \\
  & \textbf{B} &  16.80 $\,\pm\,$ 4.31 & \textbf{-0.153 $\,\pm\,$ 0.020} & \textbf{0.53 $\,\pm\,$ 0.19} & 0.043 $\,\pm\,$ 0.009 & 4.76 $\,\pm\,$ 0.49 &\textbf{ 6.037 $\,\pm\,$ 0.496 }& 65.89 $\,\pm\,$ 6.22 \\
\midrule
\multirow{3}{*}{Dynamic}
 & N & 13.43 $\,\pm\,$ 3.30 & -0.091 $\,\pm\,$ 0.025 & 56.25 $\,\pm\,$ 16.62 & 0.002 $\,\pm\,$ 0.004 & 7.26 $\,\pm\,$ 0.13 & 2.835 $\,\pm\,$ 0.437 & 45.55 $\,\pm\,$ 3.51 \\
  & P & 1.55 $\,\pm\,$ 1.80 & -0.033 $\,\pm\,$ 0.022 & 4.04 $\,\pm\,$ 9.17 & 0.023 $\,\pm\,$ 0.015 &\textbf{ 7.35 $\,\pm\,$ 0.25 }& \textbf{3.219 $\,\pm\,$ 0.911} & \textbf{33.62 $\,\pm\,$ 6.25} \\
 & \textbf{B} & 2.05 $\,\pm\,$ 3.74 & -0.024 $\,\pm\,$ 0.022 & \textbf{0.84 $\,\pm\,$ 1.92} &\textbf{ 0.017 $\,\pm\,$ 0.010} & 8.13 $\,\pm\,$ 1.32 & 5.949 $\,\pm\,$ 1.927 & 54.47 $\,\pm\,$ 4.92 \\

\bottomrule
\end{tabular}

\vspace{0.3ex}
\raggedright\footnotesize N = Baseline N (no collision avoidance); P = Baseline P (objective-based collision avoidance); \textbf{B = BarrierIK (Ours)} (CBF-based collision avoidance)
\vspace{-2em}
\end{table*}

\paragraph*{\textbf{Autonomous Evaluation Tasks and Environments}}
We evaluate BarrierIK in cluttered environments to assess both performance and generalizability. First, we perform autonomous rollouts on two complementary tasks—(i) a shelf/frame scene with tight clearances, and (ii) a dynamic-obstacle scene with prescribed obstacle motions—each designed to stress different aspects of behavior such as minimum clearance, violation time, pose tracking, and smoothness. We then test BarrierIK in a human-in-the-loop teleoperation setup. All controllers share the same non-environment objectives; BarrierIK differs only by enforcing CBF safety at the IK layer, and the same safety margin $\epsilon$.

In the \textbf{Dynamic Obstacles~(DO)} task, three cylindrical obstacles oscillate along orthogonal front–back, left–right, and vertical axes with distinct phases (capped at 0.025\,m/s), occasionally approaching the tool. The end-effector remains near center with primarily rotational adjustments, mimicking camera/tool reorientation and stressing responsiveness to moving constraints.
The \textbf{Frame / Shelf~(FS)} task contains a rectangular shelf-like frame with four large “windows” separated by vertical cylindrical bars, simulating a tool insertion task. The end-effector must pass through each opening in sequence with varying orientations. Transitions between windows are deliberately challenging: the reference trajectory lightly penetrates the frame edges, exposing whether controllers can maintain clearance without sacrificing tracking fidelity.
For each task, we specify waypoints and generate a time-parameterized end-effector trajectory. All IK solvers track the identical predefined trajectory.

We conducted a \textbf{VR-based teleoperation user study} to evaluate the proposed shared autonomy framework with a 7-DoF arm controlled via HTC Vive handheld controllers in Unity3D, as depicted in \hyperref[fig:introduction]{Figure~\ref*{fig:introduction}}.
The task was to pick three color-coded cubes and place them in a basket within clutter. The task involved picking up three color-coded cubes and placing them into a designated basket. The environment was deliberately cluttered, requiring precise navigation around obstacles to complete the task.
Ten participants (6 male, 3 female, 1 non-binary; mean age: 27.9 $\pm$ 3.14) completed trials under six configurations: (i) Baseline N without shared autonomy (N), (ii) Baseline P without shared autonomy (P), (iii) BarrierIK without shared autonomy (B), (iv) Baseline N with shared autonomy (SA-N), (v) Baseline P with shared autonomy (SA-P), and (vi) BarrierIK with shared autonomy (SA-B). Participation was voluntary, informed consent was obtained beforehand, and all data were collected and analyzed anonymously without collecting any personally identifiable information. Each participant completed five zero-shot trials per condition without prior task-specific training. A 3-minute familiarization phase preceded the trials. Condition order was randomized per user to avoid learning effects. Each trial had a 3-minute time limit and a maximum of five collisions; exceeding either threshold resulted in failure. Trials were also marked as failed if objects were dropped outside the workspace.

The complete pipeline—including simulation, shared autonomy, and IK solving—ran at 90\,Hz to match the VR headset refresh rate. Although the computational backend supported higher frequencies, rendering was the bottleneck. In 100\,Hz-capped testing, B ran at 98.2\,Hz~($\pm$ 7.91), P at 99.0\,Hz~($\pm$ 3.20), and N at a fixed 100\,Hz (due to implementation constraints). All computations ran on CPU~(Intel\textsuperscript{\textregistered} Core\texttrademark~i7-14700F); GPU acceleration was disabled for compatibility with SteamVR and JAX, which lacks stable CUDA support on Windows.

\paragraph*{\textbf{Evaluation Metrics}}
We report objective and subjective metrics across autonomous rollouts and teleoperation.

In the \textbf{autonomous rollouts setting}, we evaluate the performance of three IK solvers—Baseline~N~(N), Baseline~P (P), and BarrierIK~(B)—across two challenging tasks. All controllers share the same non-environment objectives; differences emerge only from how they handle obstacle avoidance.
We report the following metrics, averaged across 40 trials with randomized seeds:~(i) \textit{Number of Collisions}: Total link–obstacle collisions over the trajectory;~(ii) \textit{Minimum Clearance}: The smallest signed distance $\min_{\ell,o} \phi_{\ell o}(\bm{\theta})$ between any robot link~$\ell$ and any obstacle~$o$;~(iii) \textit{Violation Time Percentage}: The percentage of timesteps for which any $\phi_{\ell o} < 0$, indicating penetration;~(iv) \textit{Position Error} and \textit{Orientation Error}: The average Euclidean and rotational~(quaternion) error between the executed and reference end-effector trajectory;~(v)
\textit{Task Jerk}: The time-averaged magnitude of third-order derivatives of end-effector position, reflecting Cartesian motion smoothness;~(vi) \textit{Joint Jerk}: The time-averaged magnitude of joint-space jerk $\dddot{\bm{\theta}}(t)$, indicating low-level actuation smoothness.
For evaluation, collision checking is performed using the HPP-FCL~\cite{hppfcl}, a high-fidelity geometric proximity library capable of computing signed distances between convex capsule models. 

In the \textbf{teleoperation experiments}, participants teleoperate the robot in VR. We evaluate task efficiency and safety across the three solvers—Baseline~N, Baseline~P, and BarrierIK—each with and without shared autonomy.

As objective metric we report (i) \textit{Success rate}: percentage of trials placing all three objects without breaching time or collision limits;~(ii) \textit{Completion time}~(successful trials only), measured from first human movement $t_i$ to final placement $t_f$;~(iii) \textit{Collision count} from Unity PhysX.
Internally, solvers use JAX-based differentiable collision models, but Unity's engine provides real-time contact detection and user feedback/visualization.
We omit position/orientation tracking errors and jerk in teleoperation because human variability makes these low-level metrics less diagnostic for comparison.

To assess perceived usability based on subjective metrics, we administered a modified NASA Task Load Index (TLX)~\cite{hart_1988_nasatlx} after each configuration, in which all dimensions were inverted so that higher scores indicate more favorable outcomes. The inverted scales included physical~(I-PD), temporal~(I-TD), and mental demand~(I-MD), perceived effort~(I-EFF), frustration~(I-FL), and perceived performance~(I-PER). We further extended the TLX with dimensions specific to shared autonomy, namely perceived control level~(CL), assistance~(AL), and safety~(SL). These ratings were visualized as polar plots, where larger areas represent a more favorable user experience. Finally, participants provided a ranked list of the six configurations from most to least preferred after completing all trials.


\section{Results}
\label{sec:result}
\captionsetup[subfigure]{justification=justified,singlelinecheck=false}
\begin{figure}[t]
    \centering
    \includegraphics[width=\linewidth]{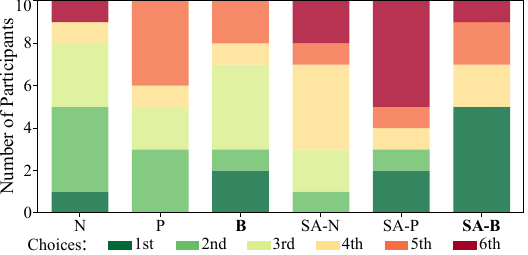}
    \vspace{-1.5em}
    \caption{The stacked bar chart illustrates the number of participants who ranked each controller configuration from 1st to 6th choice. Each color segment within a bar represents the count of users selecting that configuration at the respective rank.}
    \label{fig:subjective:user}
    \vspace{-2em}
\end{figure}

This section presents our experimental findings across autonomous and teleoperation settings. We evaluate BarrierIK against baseline methods using standard safety and performance metrics, and report both quantitative results and user study outcomes.

\paragraph*{\textbf{Autonomous Trajectory - Evaluation Results}}
Across both scenes, \textbf{B} achieves markedly lower \emph{violation time} and less negative \emph{minimum clearance} than \textbf{P}, indicating that when contacts do occur they are \emph{shorter and shallower}. 
The residual violation time for \textbf{B} results from discrete-time position updates and SLSQP’s local linearization, which can lead to transient numerical penetrations between control steps.
Practically, the controller does not linger near the boundary but returns to the safe set promptly—i.e., it \emph{recovers} from adverse states rather than getting \emph{stuck} at the constraint. 

\begin{figure}[t]
\centering
\includegraphics[width=1\linewidth]{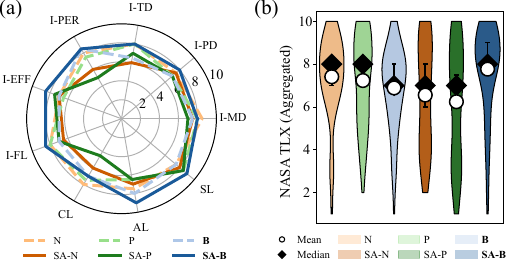}
\vspace{-1.75em}
\caption{Subjective workload and usability across teleoperation configurations.
\textbf{(a)} Polar (“radar”) plot of modified NASA-TLX dimensions—physical demand (I-PD), temporal demand (I-TD), mental demand (I-MD), performance (I-PER), effort (I-EFF), frustration (I-FL), control level (CL), assistance level (AL), and safety level (SL)—with “I-” denoting inverted scales so that larger values correspond to more favorable assessments. \textbf{(b)} Violin plots of participants' aggregated NASA-TLX scores for each configuration (mean $\circ$, median $\blacklozenge$), where lower workload (higher support) appears toward the top of the scale.}
\label{fig:subjective:tlx}
\vspace{-2em}
\end{figure}

\paragraph*{\textbf{Teleoperation - Subjective Results}}
Participants split into two clear archetypes: a group favoring \textbf{N} (no assistance) and another preferring \textbf{SA-B} (shared autonomy with BarrierIK) (Fig.~\ref{fig:subjective:user}). These configurations sit at opposite ends of the autonomy spectrum, yet both deliver favorable experiences, indicating that perceived quality depends as much on \emph{how} assistance interacts with the operator as on \emph{how much} assistance is provided.

\begin{figure*}[t!]
\centering
\includegraphics[width=1\linewidth]{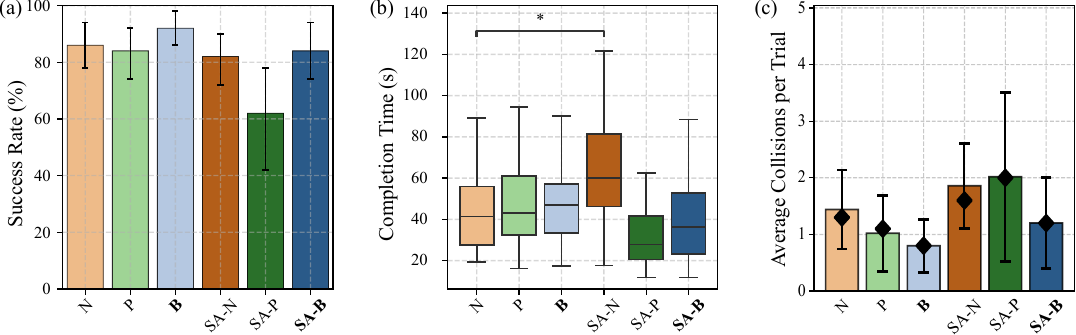}
\vspace{-1.5em}
\caption{Objective evaluation across six system configurations. Subplots \textbf{(a)}–\textbf{(c)} report performance and safety metrics: success rate (higher is better), task completion time (lower is better, measured only for successful trials), number of collisions (lower is better).  $\blacklozenge$ indicates the median,  asterisks indicate statistical significance: $^*$~$p<.05$, $^{**}$~$p<.01$, $^{***}$~$p<.001$.}
\label{fig:objective}
\vspace{-2em}
\end{figure*}

On the inverted NASA–TLX radar (Fig.~\ref{fig:subjective:tlx}a), \textbf{N} and \textbf{SA-B} cover the largest areas: \emph{lower perceived mental, physical, temporal demand, effort,} and \emph{frustration}, together with higher \emph{control level}, \emph{assistance level}, and \emph{self-reported safety level}. Notably, \textbf{SA-B} concentrates responses near the favorable region, suggesting that constraint-aware assistance can reduce workload without eroding agency, consistent with high CL alongside high AL and SL. In contrast, intermediate shared-autonomy modes (\textbf{SA-N}, \textbf{SA-P}) exhibit smaller, more dispersed footprints, reflecting mixed impressions: when assistance deviates from user intent, operators report reduced control and lower perceived support.

To summarize across dimensions, we aggregated inverted workload scores with CL/AL/SL into a composite measure (Fig.~\ref{fig:subjective:tlx}). \textbf{SA-B} achieved the highest average score, followed by \textbf{N}. The \textbf{SA-B} distribution also appears tighter, suggesting more consistent relief in perceived workload across participants. With $n{=}10$, a Friedman test with Nemenyi post-hoc found no significant pairwise differences (all $p{>}0.05$), and therefore the observed patterns should be interpreted as indicative trends rather than conclusive effects. Nevertheless, the results indicate a coherent tendency: operators either prefer the transparency and predictability of \textbf{N} or the consistent, constraint-aware guidance of \textbf{SA-B}. 


We interpret the results as follows: Participants accept help if the system prioritizes safety and they can retain control. Employing CBF at the IK layer guides actions to morph within safe and feasible regions rather than overruling the operator. Accordingly, \textbf{SA-B} scores better on CL (control), AL (assistance), and SL (safety) than \textbf{SA-N}/\textbf{SA-P}.

\paragraph*{\textbf{Teleoperation - Objective Results}}
For \emph{success rate}, results cluster around $\sim$80\% for most configurations; \textbf{SA-P} is the exception with a lower average success rate and higher variability (Fig.~\ref{fig:objective}a). The non-assisted \textbf{B} configuration trends toward a higher median success rate with a tighter interquartile range; however, pairwise differences are not statistically significant at our sample size.
In terms of \emph{completion time} (successful trials only), \textbf{SA-P} appears faster on successful trials (Fig.~\ref{fig:objective}b), but this reflects the exclusion of failed runs rather than an inherent efficiency advantage.
For \emph{collisions}, unstructured assistance tends to increase contacts in cluttered environments. \textbf{SA-N} and \textbf{SA-P} show higher collision counts than their non-SA counterparts (Fig.~\ref{fig:objective}c). Among shared-autonomy modes, \textbf{SA-B} reduces collisions, while \textbf{B} (no SA) remains best overall on raw collision count.

\section{Conclusion and Future Work}
\label{sec:conclusion}
We presented a hierarchical shared-autonomy framework that separates goal-directed guidance from IK-level safety enforcement. The safety layer, \textbf{BarrierIK (B)}, integrates control barrier functions with an optimization-based IK to systematically address safety \emph{after} blending human and autonomous inputs. Our evaluation comprised (i) autonomous rollouts in representative cluttered scenes and (ii) a VR teleoperation study with and without shared autonomy. In autonomous rollouts, \textbf{B} consistently reduced \emph{violation time} and yielded \emph{less-negative minimum clearance} compared to the cost-based baseline \textbf{P}. These findings indicate shorter, shallower boundary contacts and more reliable recovery to the safe set. In teleoperation, assistance \emph{without} an IK-level safety filter~(\textbf{SA-N}, \textbf{SA-P}) tended to degrade outcomes in clutter~(e.g., lower success, higher collisions), whereas our method \textbf{SA-B} recovered safety and maintained user acceptance, producing the most balanced overall profile when considering both subjective and objective criteria. We refrain from pairwise significance claims on collision counts at our sample size, and interpret these patterns as consistent trends that align with our predefined hypothesis that projecting blended commands from CBFs into the safety set preserves task performance while increasing constraint satisfaction and maintaining user satisfaction. 
\paragraph*{\textbf{Limitations}} Treating safety as a hard constraint can induce detours; in our study, this appears as a higher average joint jerk for \textbf{SA-B}. Although violation time and clearance are reduced, this safety–smoothness trade-off may be undesirable for contact-sensitive tasks. Our CBF implementation is discrete and, in dynamic scenes, does not incorporate obstacle-velocity estimates; the resulting safe set can be conservative when obstacles move. Design choices for the barrier~(e.g., class-$\mathcal{K}$ functions and shaping) materially influence intervention behavior and were not tuned per-user or per-scene. Finally, while CBFs are typically posed in velocity/torque space, our teleoperation stack runs in position space; we therefore approximate a task-space velocity via a temporal-difference between the current end-effector pose and the shared-autonomy target. This myopic, step-wise approximation enables IK layer CBF enforcement at real-time teleoperation rates but may miss longer-horizon structure.

\paragraph*{\textbf{Future work}} We plan to deploy on physical hardware and directly measure contact dwell to quantify recovery, incorporate velocity observers to update dynamic CBF constraints online, and study the impact of class-$\mathcal{K}$ choices and adaptive shaping on user acceptance and intervention rates with a more extensive user study. Overall, the results show that an optimization-based IK solver can be equipped with CBF constraints to provide \emph{myopic safety awareness} at teleoperation rates, yielding intent-preserving, constraint-aware assistance in cluttered environments.





% ...


% ...


\bibliographystyle{styles/IEEEtran}
\bibliography{bib/references}

\end{document}